\documentclass[a4paper, 10pt, conference]{ieeeconf}      

\IEEEoverridecommandlockouts                              
\usepackage{graphicx} 
\usepackage{xcolor}
\usepackage{amssymb}
\usepackage{float}
\usepackage{flushend}

\title{\LARGE \bf
Robust and fast generation of top and side grasps for unknown objects
}

\author{Brice Denoun$^{1,2}$, Beatriz Leon$^{2}$, Claudio Zito$^{3}$, Rustam Stolkin$^3$, Lorenzo Jamone$^{1}$ and Miles Hansard$^{1}$.
\thanks{$^{1}$ School of Electronic Engineering and Computer Science, Queen Mary University of London, United Kingdom {\tt\small \{b.d.denoun, l.jamone, m.hansard\}@qmul.ac.uk}}
\thanks{$^{2}$ The Shadow Robot Company, London, United Kingdom}
\thanks{$^{3}$ Extreme Robotics Lab, School of Metallurgy and Materials, University of Birmingham, United Kingdom {\tt\small \{C.Zito, R.Stolkin\}@bham.ac.uk}}
}%

\begin{document}

\maketitle
\thispagestyle{empty}
\pagestyle{empty}


\section*{Introduction}
Grasping is a skill which humans use every day to interact with their environment. Interestingly, the grasps depend on the pose and shape of the objects, but also on their intended use in each specific situation. Being able to replicate this human skill with robots is an active research area that has been studied for decades, and has multiple applications, both in industry and in less structured environments, such as hazardous environments, public spaces and homes.\\
Finding good grasp \emph{configurations} for objects is usually referred as Grasp Pose Detection (GPD). Many recent studies addressing this problem can be categorised in two families, geometry-based and Machine Learning-based methods. The first approach doesn't need training; it relies on geometric properties extracted from the scene, and on heuristics observed when humans manipulate objects~\cite{suzuki2016grasping, kundu2018novel}. The second leverages Machine Learning (ML) techniques to learn how to predict robust grasp poses~\cite{kopicki_2015,zito_2019,lenz_2015, mahler2017dex}. These methods usually do not need feature engineering but require a considerable amount of data and time to train.\\ 
With the emergence of cheap vision devices (RGB and \mbox{RGB-D} sensors), studies carried out so far focus mainly on using visual information in order to solve the aforementioned problem. As a result, different strategies have been proposed to extract useful information about the scene. Some methods use multiple fixed sensors, others opt for a camera device mounted on the wrist, which enables visual servoing to refine the grasp pose. Finally, other methods rely on a single fixed view, and predict how the robot should grasp the object with a limited amount of information. \\

In this work, we present a geometry-based grasping algorithm that is capable of efficiently generating both top and side grasps for unknown objects, using a single view RGB-D camera, and of selecting the most promising one. We demonstrate the effectiveness of our approach on a picking scenario on a real robot platform. Our approach has shown to be more reliable than another recent geometry-based method considered as baseline~\cite{suzuki2016grasping} in terms of grasp stability, by increasing the successful grasp attempts by a factor of six.

\section*{Grasping method}
Our method determines two (pregrasp and grasp) 6D poses from a point cloud captured from a single RGB-D sensor. In this section, we summarise the important steps from data acquisition to the grasp pose generation. 
Given a raw input point cloud, all unreachable structures are cropped. We then use Random Sample Consensus (RANSAC)~\cite{fischler1981random} to detect and fit the dominant plane in the remaining data. The outlying points, with respect to this plane, are assumed to constitute the object of interest.
\begin{figure}[t]
	\centering
	\includegraphics[width=0.35\textwidth]{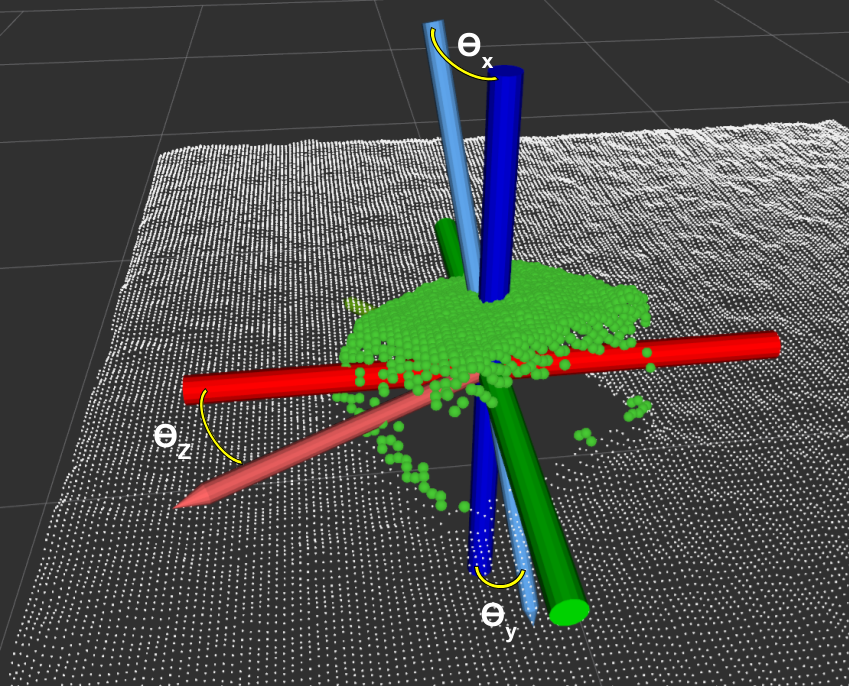}
    \caption{\emph{To compute the required position and orientation of the gripper, the centroid and the principal elongation axes of the object point cloud are used, respectively.}}
    \label{fig1}
\end{figure}
The next step is to infer the centroid $\mathbf{X}_c$ of the object, without identifying its shape. This is motivated by the fact that for pick and place related tasks, humans will naturally grasp the object around its centre of mass. Determining this physical property from a point cloud is still a challenging problem in vision that we do not want to tackle. We simplify this problem by assuming both points are equivalent. Instead of directly using the average of the extracted point cloud (as in the baseline~\cite{suzuki2016grasping}), we are using the knowledge brought by the previous plane segmentation. It allows to avoid having a centroid predicted in the top part of the object (which can lead to poor quality grasps) depending on the RGB-D pose. For this purpose we correct the centroid's $z$-coordinate with the plane height.
In order to determine the pose of the end-effector, we need to estimate the orientation of the object. We do this by Principal Component Analysis (PCA), which determines the object's three principal axes $\mathbf{u}_i$, with $i \in \{1,2,3\}$, along with the corresponding eigenvalues $\lambda_1 \ge \lambda_2 \ge \lambda_3$. Using the latter, we can distinguish two scenarios: $a)$ the object is standing upright (principal axis is pointing along $z$-axis) or $b)$ it is lying down on the surface (principal axis perpendicular to the $z$-axis).

As illustrated on Figure \ref{fig1}, the computation of Euler angles, $\theta_x, \theta_y, \theta_z$, depends on the estimated orientation, and our method automatically adapts and generates both side and top grasps, according to the object's centroid and height.
Finally we define the pre-grasp and grasp position as \mbox{$\mathbf{X}_c + h \mathbf{u}_i$}. In scenario $a)$, $\mathbf{u}_i$ corresponds to the axis pointing mainly upward. In scenario $b)$, $u_i$ is chosen to be the second principal axis. For the pre-grasp position, $h$ is defined by the user to be how far the end-effector should be before approaching the object. For the grasp position, $h$ is equivalent to the length of the manipulator's fingers so that when closing it squeezes the surface of the object around its centroid.

\section*{Preliminary results}
In order to demonstrate the robustness of our method, we used a UR5 robot arm, an \texttt{EZGripper}, an under-actuated two finger gripper and a Kinect v2. The latter was fixed perpendicularly $1.72$m above a $60 \times 60$cm table. All the code is implemented in \texttt{Python} and was executed on a single laptop (Intel® Core™ i7-8750H CPU @ 2.20GHz $\times$ 12 with 16 Go RAM). We rely on the \texttt{MoveIt!} framework~\cite{coleman2014reducing} for the motion planning task.\\
The first results used to evaluate our method compare the robustness of the generated grasps compared to the method proposed by~\cite{suzuki2016grasping}. So far, we have run both algorithms on five objects (a cardboard tube, a screwdriver, a pair of thick plastic gloves, duct tape and one of the adversarial objects introduced in~\cite{mahler2017dex}) in three poses with five repetitions each. Most of these objects are part of a wider set of objects established within the National Centre for Nuclear Robotics (NCNR), which gathers challenging objects to grasp in order to compare methods for a variety of tasks.\\
\begin{table}[H]
\begin{tabular}{|c|c|c|c|}
\hline
  & Failed attempt & Unstable grasp & Dropped object \\
\hline
Baseline~\cite{suzuki2016grasping} & 22.7\% & 6.7\% & 10.7\% \\
Our method & $\mathbf{4\%}$ & $\mathbf{0\%}$ & $\mathbf{1.3\%}$ \\
\hline
\end{tabular}
\caption{Failure rates on a 75 grasp trials}
\label{tab:results}
\end{table}
Once the gripper is closed, the arm is moved to a specific location and an empirical stability check is performed. If the gripper is empty before starting the stability check, then the grasp is recorded as failed attempt. The arm is then lifted and left static for an arbitrary amount of time ($3$s) to test if the grasp is gravity-resistant. If it not, it is recorded as unstable. The final part of the check is an automatic and reproducible pre-programmed shake of the arm. If the object falls, then this is also recorded (please refer to Table \ref{tab:results}). To the best of our knowledge this grasp stability evaluation has never been performed in previous works.
Our method has approximately six times fewer failed attempts when grasping the objects, with respect to the baseline method. For the latter, the failed attempts are mainly due to the fact that the model cannot inherently generate grasps for standing objects (a problem that we address by side grasps). The failed grasp predictions of our method all occurred when generating grasps for the plastic gloves, especially when they were placed in a \emph{flat} configuration (and thus easily confused with the table plane).
Our method also addresses the issue of grasps that let the object fall during the stability check. These results seem to point out that generating side grasps and refining the way the centroid is estimated improves the robustness of the GPD.

\section*{Future work}
The proposed method could be improved by refining how the centroid of the object is estimated, as it is likely to fail on highly non-convex shapes (e.g.~pliers). In addition, improving the plane detection would allow us to detect flat and small objects such as dice or gloves in more challenging poses.
The next step will be to carry out more in-depth experiments (by using the NCNR set with more repetitions and more poses) to evaluate our algorithm, and compare it with Machine Learning based methods.
Finally, we would like to extend this geometry-based approach to grasping in cluttered environments.

\section*{Conclusion}
We propose in this work a fast method that uses primitive geometric features extracted from a partial point cloud to generate top and side grasps. First results show that our algorithm generates significantly more robust grasps in different conditions (p-values $< 10^{-3}$).\\ 
In addition, this method can be transferred to a multi-fingered hand since it does not rely on any CAD model of the robot.

\section*{Acknowledgements}
This work was supported by the EPSRC UK (project NCNR, National Centre for Nuclear
Robotics, EP/R02572X/1) and by The Shadow Robot Company.


\addtolength{\textheight}{-12cm}   





\bibliographystyle{abbrv}
\bibliography{mybib}

\begin{thebibliography}{1}

\bibitem{coleman2014reducing}
D.~Coleman, I.~Sucan, S.~Chitta, and N.~Correll.
\newblock Reducing the barrier to entry of complex robotic software: a moveit!
  case study.
\newblock {\em arXiv preprint arXiv:1404.3785}, 2014.

\bibitem{fischler1981random}
M.~A. Fischler and R.~C. Bolles.
\newblock Random sample consensus: a paradigm for model fitting with
  applications to image analysis and automated cartography.
\newblock {\em Communications of the ACM}, 24(6):381--395, 1981.

\bibitem{kopicki_2015}
M.~Kopicki, R.~Detry, M.~Adjigble, R.~Stolkin, A.~Leonardis, and J.~L. Wyatt.
\newblock One-shot learning and generation of dexterous grasps for novel
  objects.
\newblock {\em The International Journal of Robotics Research}, 35(8):959--976,
  2015.

\bibitem{kundu2018novel}
O.~Kundu and S.~Kumar.
\newblock A novel geometry-based algorithm for robust grasping in extreme
  clutter environment.
\newblock {\em arXiv preprint arXiv:1807.10548}, 2018.

\bibitem{lenz_2015}
I.~Lenz, H.~Lee, and A.~Saxena.
\newblock Deep learning for detecting robotic grasps.
\newblock {\em International Journal of Robotics}, 34:705--724, 2015.

\bibitem{mahler2017dex}
J.~Mahler, J.~Liang, S.~Niyaz, M.~Laskey, R.~Doan, X.~Liu, J.~A. Ojea, and
  K.~Goldberg.
\newblock Dex-net 2.0: Deep learning to plan robust grasps with synthetic point
  clouds and analytic grasp metrics.
\newblock {\em arXiv preprint arXiv:1703.09312}, 2017.

\bibitem{suzuki2016grasping}
T.~Suzuki and T.~Oka.
\newblock Grasping of unknown objects on a planar surface using a single depth
  image.
\newblock In {\em Advanced Intelligent Mechatronics (AIM), 2016 IEEE
  International Conference on}, pages 572--577. IEEE, 2016.

\bibitem{zito_2019}
C.~Zito, V.~Ortenzi, M.~Adjigble, M.~Kopicki, R.~Stolkin, and J.~L. Wyatt.
\newblock Hypothesis-based belief planning for dexterous grasping.
\newblock {\em arXiv preprint arXiv:1903.05517 [cs.RO] (cs.AI)}, 2019.

\end{thebibliography}

\end{document}